\documentclass[letterpaper]{article} % DO NOT CHANGE THIS
\usepackage{aaai23}  % DO NOT CHANGE THIS
\usepackage{times}  % DO NOT CHANGE THIS
\usepackage{helvet}  % DO NOT CHANGE THIS
\usepackage{courier}  % DO NOT CHANGE THIS
\usepackage[hyphens]{url}  % DO NOT CHANGE THIS
\usepackage{graphicx} % DO NOT CHANGE THIS
\usepackage{natbib}  % DO NOT CHANGE THIS AND DO NOT ADD ANY OPTIONS TO IT
\usepackage{caption} % DO NOT CHANGE THIS AND DO NOT ADD ANY OPTIONS TO IT
\usepackage{algorithm}
\usepackage{algorithmic}
\usepackage{amsmath}
\usepackage{amssymb}
\usepackage{threeparttable}
\usepackage{booktabs}
\usepackage{multirow}
\usepackage{subfigure}

\usepackage{newfloat}
\usepackage{listings}
\DeclareCaptionStyle{ruled}{labelfont=normalfont,labelsep=colon,strut=off} % DO NOT CHANGE THIS
\floatstyle{ruled}
\newfloat{listing}{tb}{lst}{}
\floatname{listing}{Listing}
\nocopyright 
\title{LKD-Net: Large Kernel Convolution Network for Single Image Dehazing}
\author {
    Pinjun Luo \textsuperscript{\rm 1}
    Guoqiang Xiao \textsuperscript{\rm 1}
    Xinbo Gao \textsuperscript{\rm 2}
    Song Wu \textsuperscript{\rm 1}\thanks{Corresponding author}
}
\affiliations {
    \textsuperscript{\rm 1} College of Computer and Information Science, Southwest University\\
    \textsuperscript{\rm 2} Chongqing University of Posts and Telecommunications\\
    zhongjiuhuan@email.swu.edu.cn,  gqxiao@swu.edu.cn, gaoxb@cqupt.edu.cn, songwuswu@swu.edu.cn
}

\begin{document}

\maketitle

\begin{abstract}
The deep convolutional neural networks (CNNs)-based single image dehazing methods have achieved significant success. The previous methods are devoted to improving the network's performance by increasing the network's depth and width. The current methods focus on increasing the convolutional kernel size to enhance its performance by benefiting from the larger receptive field. However, directly increasing the size of the convolutional kernel introduces a massive amount of computational overhead and parameters. Thus, a novel Large Kernel Convolution Dehaze Block (LKD Block) consisting of the Decomposition deep-wise Large Kernel Convolution Block (DLKCB) and the Channel Enhanced Feed-forward Network (CEFN) is devised in this paper. The designed DLKCB can split the deep-wise large kernel convolution into a smaller depth-wise convolution and a depth-wise dilated convolution without introducing massive parameters and computational overhead. Meanwhile, the designed CEFN incorporates a channel attention mechanism into Feed-forward Network to exploit significant channels and enhance robustness. By combining multiple LKD Blocks and Up-Down sampling modules, the Large Kernel Convolution Dehaze Network (LKD-Net) is conducted. The evaluation results demonstrate the effectiveness of the designed DLKCB and CEFN, and our LKD-Net outperforms the state-of-the-art. On the SOTS indoor dataset, our LKD-Net dramatically outperforms the Transformer-based method Dehamer with only 1.79\% \#Param and 48.9\% FLOPs. The source code of our LKD-Net is available at https://github.com/SWU-CS-MediaLab/LKD-Net.   
\end{abstract}

\section{Introduction}
Single image dehazing is an ill-posed problem in computer vision. Haze can block objects in the image, which seriously causes information to degrade (e.g., color distortion, visibility decrease). Such information degradation can result in severe consequences in some scenes, such as autonomous driving, and adversarial attacking. Therefore, how to remove haze from a single image becomes a challenge in computer vision.

\begin{figure}[t]
	\centering
	\includegraphics[width=0.95\columnwidth]{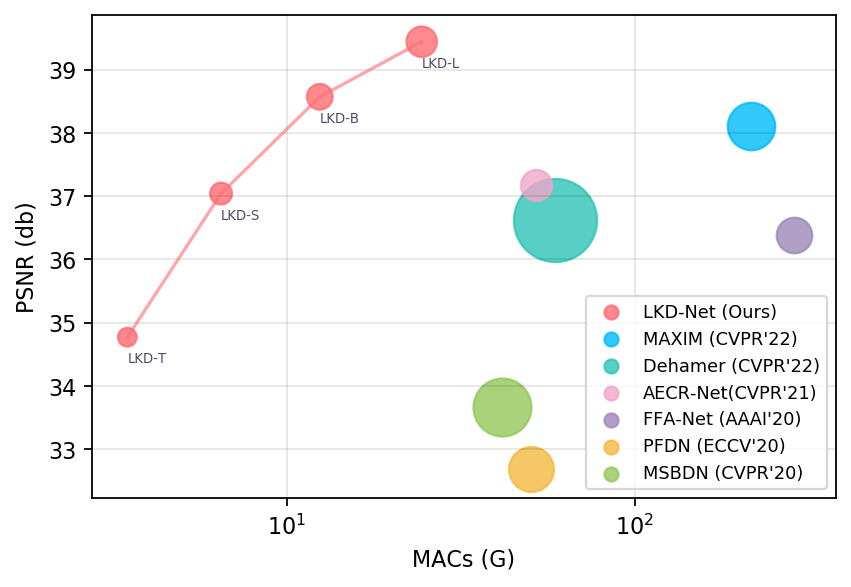} % Reduce the figure size so that it is slightly narrower than the column. Don't use precise values for figure width.This setup will avoid overfull boxes. 
	\caption{Results of different dehazing methods on the SOTS indoor set. Comparing the performance of recent methods MAXIM, Dehamer, AECR-Net, FFA-Net, PFDN, MSBDN and our LKD-Net. Network parameters, indicated by radiuses of circles, and FLOPs are shown with a logarithmic axis. Note that our LKD-Net obtains higher PSNR while having less model complexity.}
	\label{fig1}
	
\end{figure}

The goal of single image dehazing is to estimate the latent haze-free image from the observed hazy image. Early image dehazing methods are mainly based on the atmosphere scattering model \cite{NarasimhanN00}, \cite{NarasimhanN02}, which is formulated as:
\begin{equation}
\label{eqn:asm}
    I(x) = J(x) t(x)+A (1-t(x)),
\end{equation}

Where $I(z)$ is the hazy image, $J(x)$ is the haze-free image, $A$ is the medium global atmosphere light, and $t(x)$ is the medium transmission map, the formula of $t(x)$ can be expressed as:
\begin{equation}
\label{eqn:asmt}
   t(x) = \mathrm{e}^{-\beta d(x)},
\end{equation}

Where $\beta$ is the scattering coefficient of the atmosphere and $d(x)$ is the scene depth. According to the formulation \ref{eqn:asm} and \ref{eqn:asmt}, if we estimate the global atmosphere light and medium transmission map for a hazy image, we can get the latent haze-free image from the observed hazy image. However, this prior-based model is susceptible to different scenario priors and therefore has poor robustness and is gradually being phased out.

The recent deep convolutional neural networks \cite{krizhevsky2012imagenet} (CNNs) have witnessed success in single image dehazing, and the deep learning-based image dehazing methods can be divided into two main categories. The first category of CNNs-based methods mainly focused on increasing the depth and width of the networks or designing a large kernel convolution to improve the performance of image dehazing \cite{van}, \cite{2}, \cite{3}, \cite{4}. 
Introducing large kernels to CNNs can enlarge the receptive field and capture more structured information in the learned latent domain space. However, expanding the size of the convolutional kernel results in more computational overhead and parameters, because of the quadratic computational complexity with the kernel size \cite{repLK}. The second category is the Vision Transformers (a.k.a., ViTs) based methods \cite{guo2022image}, \cite{zhao2021hybrid}. Since the operation of the large kernel in the multi-head self-attention (MHSA) mechanism, the MHSA in ViTs can integrate information from large receptive fields and model long-range dependencies. It can be noted that the large kernel operation in both CNNs-based and Vision Transformer-based methods can effectively build a large effective receptive field to improve performance. However,  the large-kernel operation is computationally expensive because the kernel size quadratically increases the number of parameters and floating point operations (FLOPs). Moreover, it is also found that CNNs-based and Vision Transformer-based image dehazing methods handle the channels in Feed-forward Network (FN) equally important. However, various channel features in FN have additional weighted information. If we treat them equally, the FN will spend many resources on the unnecessary computation of unimportant channels and significantly reduce the efficiency of the network optimization.

Thus, a novel Large Kernel Convolution Dehaze Block (LKD Block) is designed in this paper to address the above-mentioned limitations. Specifically, motivated by the convolution decomposition mechanism \cite{van}, \cite{restormer}, a decomposition of the Large Kernel Convolution Block (DLKCB) is designed to replace the multi-head self-attention (MHSA) in ViTs. DLKCB increases the effective receptive field and builds long-range information between features without adding more computational overhead and parameters by decomposing the traditional large depth-wise kernel convolution into a combination of a small depth-wise convolution and a depth-wise dilated convolution. Furthermore,  the Channel Enhanced Feed-forward Network (CEFN) is designed, which integrates a Channel Attention mechanism \cite{SENet} to the conventional FN to improve the efficiency of the network optimization by exploiting significant and critical channels in FN. By combining DLKCB and CEFN, our Large  Kernel Convolution Dehaze Block (LKD Block) is conducted, which can be treated as a plug-in and added to the deep architecture of CNNs and ViTs for both high-level and low-level computer vision tasks. This paper adds the LKD Block to the multiple U-Net-like dehazing networks for high-performance Single Image Dehazing (denoted as LKD-Net).  

To the best of our knowledge, our LKD-Net is the first method to employ the large depth-wise kernel convolution in the task of single image dehazing. Figure \ref{fig1} shows the comparison between LKD-Net and the compared state-of-the-art methods on the SOTS indoor set. We can note that our LKD-Net outperforms the previous Swin Transformer-based \cite{guo2022image} on the SOTS \cite{RESIDE} indoor set with significantly fewer computational overhead and parameters. Moreover, LKD-Net has the same scale-up capability as the Transformer-based method. The main contributions of our LKD-Net can be summarized as follows:

\begin{figure*}[t]
	
	\centering
	\includegraphics[scale=0.48]{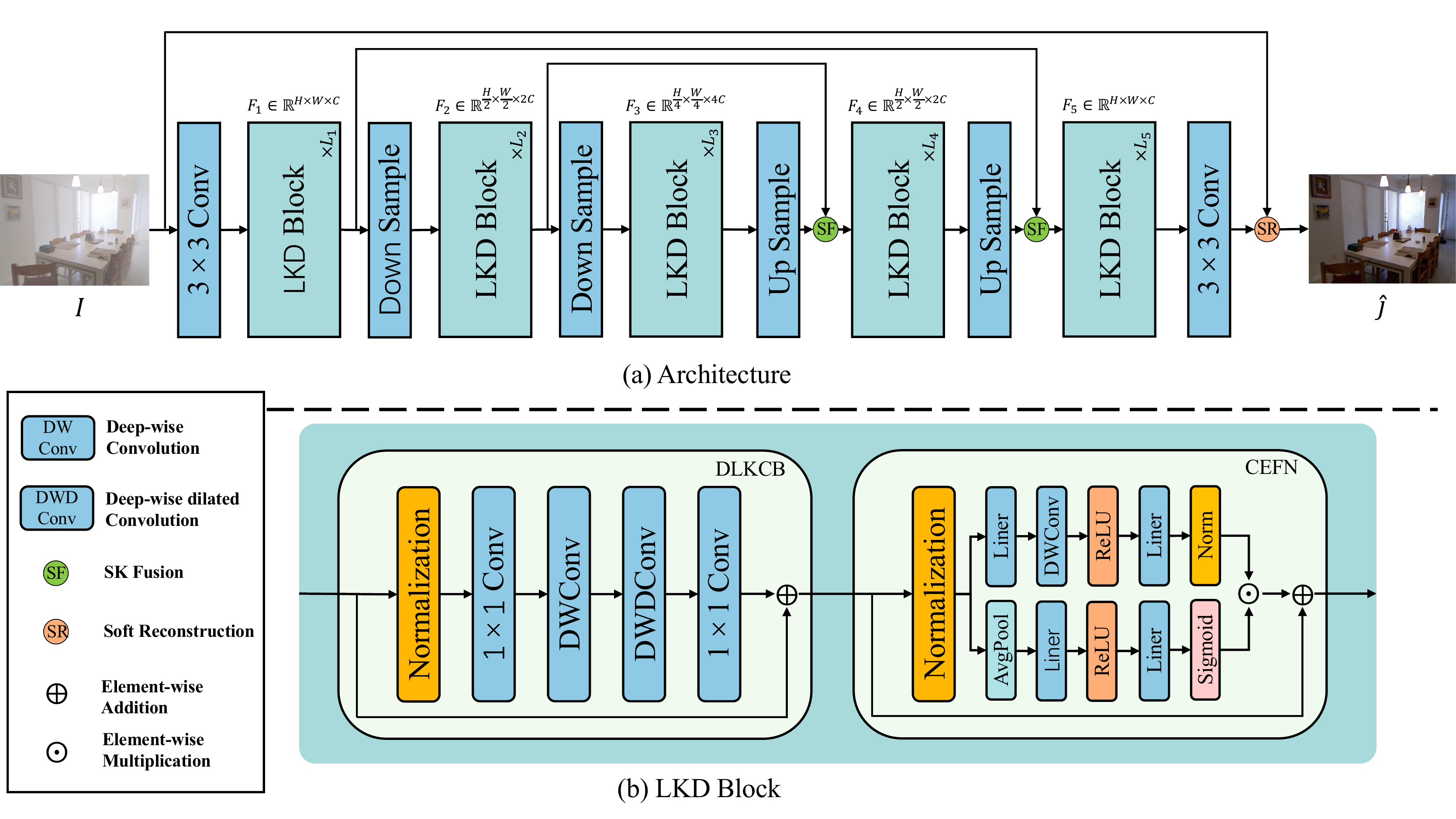} % Reduce the figure size so that it is slightly narrower than the column.
	\caption{(a) Large Kernel Convolution Network (LKD-Net) architecture. (b) The architecture of LKD Block. }
	\label{fig:flowchart}
	
\end{figure*}

%Overall, our contributions are summarized as follows:
\begin{itemize}
\item We proposed the LKD-Net, a high efficient end-to-end multiple U-Net-like deep architecture for single image dehazing. LKD-Net outperforms the state-of-the-art methods by using significantly fewer parameters and lower computational overhead.
\item We designed the Large Kernel Convolution Dehaze Block (LKD Block), which can be used as a plug-in module to enhance the performance of both CNNs and Transformers architecture. Meanwhile, the LKD Block is more efficient and effective for the single image dehazing task than Transformer-based methods.
\item We designed the Decomposition Large Kernel Convolution Block (DLKCB), which decomposes the large depth-wise convolution into a small depth-wise convolution and a depth-wise dilated convolution to increase the effective receptive field without increasing massive parameters and computational overhead. 
\item We designed the Channel Enhanced Feed-forward Network (CEFN), which can effectively explore and integrate the channels with more critical information in FN, further improving the robustness and efficiency of the network optimization.
\end{itemize}

\section{Related Work}
Single image dehazing methods can be classified as prior-based, CNNs-based, and Transformer-based methods.

\textbf{Prior-based Image Dehazing Methods.} These methods mainly depend on the atmosphere scattering model and the handcraft priors. DCP \cite{DCP} proposed to use the dark channel prior to estimate the medium transmission map. \cite{DBLP:conf/bmvc/ZhuMS14} proposed color attenuation prior by using a linear model to model the scene depth of a hazy image. \cite{DBLP:conf/cvpr/BermanTA16} proposed a dehazing algorithm using a Non-Local prior that hundreds of different color clusters well approximate the color of a haze-free image in RGB space. However, these prior-based methods all have the disadvantage of being susceptible to different scenarios resulting in less robustness.

\textbf{CNNs-based Image Dehazing Methods.} Since the high semantic abstraction capability of CNNs \cite{AlexNet}, DehazeNet \cite{DehazeNet} employed CNNs for single image dehazing by using CNNs to estimate the medium transmission map and restore the hazy images by the atmosphere scattering model. FFA-Net \cite{2} utilized CNNs to build a feature fusion attention mechanism to process different channel information and pixel information flexibly. AECR-Net \cite{3} has improved the performance on FFA-Net by using downsample and contrastive learning. However, these CNNs-based methods focused only on increasing the depth and width of the network instead of on the kernel size. This is mainly because directly expanding the size of the convolutional kernel results in introducing more computational overhead and parameters, considering the quadratic computational complexity with the size of the kernel.

\textbf{Transformer-based Image Dehazing Methods.} Since \cite{ViT} introduced Transformer to computer vision, ViTs have surpassed traditional CNNs-based methods in various tasks, including the single image task. For instance, \cite{dehazeformer} proposed the Dehazeformer, which uses the Swin Transfomer as the backbone, surpasses all previous CNNs-based methods by a large margin on the SOTS \cite{RESIDE} dataset. Some methods \cite{liu2022convnet}, \cite{repLK} transformed CNNs into Transformer-like CNNs architectures, and have also yielded promising results in various fields. However, these methods usually spend a lot of resources on processing tokens and ignore that different channels have entirely different information weights in FN, resulting in a less efficient network.

\section{Method}
% In this section, we first introduce the overall structure of our LKD-Net. Then, we will introduce the LKD Block, which is one most important components of our proposed LKD-Net. The LKD Block is constructed by combining the Decomposition Large Kernel Convolution Block (DLKCB) and the Channel Enhanced Feed-forward Network (CEFN). Finally, we will describe DLKCB and CEFN, respectively.

\subsection{Overall Architecture}
Our main goal is to develop an efficient deep model which can restore the hazy observation images to hazy-free images. As presented in Figure\ref{fig:flowchart} (a), the LKD-Net is a U-Net-like architecture, a multi-scale hierarchical framework containing multiple LKD Blocks, which has the significant advantage of improving performance while maintaining no additional computational cost. The downsample layer uses non-intersecting convolution to divide the image into small patches with non-intersecting and increase the number of channels. The upsample layer uses Pixelunshuffle \cite{ledig2017photo} to aggregate the corresponding downsample layer patches and reduce the number of channels. The SK Fusion \cite{dehazeformer} is used to replace the concatenation fusion layer, which uses a channel attention mechanism to fuse the features of different branches. The Soft Reconstruction \cite{dehazeformer} layer is used to replace global residual learning, which introduces a weakly constrained prior to global residual learning resulting in better network performance.

Feed a hazy image $I\in{\mathbb R^{H\times{W\times{3}}}}$ into the LKD-Net, it restores its corresponding hazy-free image $\hat{J}\in{\mathbb R^{H\times{W\times{3}}}}$. Specifically, the hazy input image $I$ is first passed through a $3\times3$ convolution to obtain a low-level feature embedding $F_1\in{\mathbb R^{H\times{W\times{C}}}}$; where $H$ is the height of the image, $W$ is the width of the image, and $C$ is the number of channels. Subsequently, the shallow features $F_1$ are sent to a 3-layer symmetric encoder-decoder architecture to extract the depth features, and finally, the output $F_5\in{\mathbb R^{H\times{W\times{C}}}}$ is obtained. The SK Fusion is used to help the network recover images better by fusing features from the same encoder and decoder layers. Moreover, for the obtained $F_5$, a $3\times3$ convolution with the Soft Reconstruction is operated to obtain a hazy-free image $\hat{J}$.

\subsection{Large Kernel Convolution Dehaze Block}
As shown in Figure \ref{fig:flowchart} (b), our LKD Block mainly contains two modules, the designed DLKCB and CEFN. The DLKCB is used to process the information of spatial dimension, which benefits the network to retain more spatial structure information by increasing the effective receptive field. The CEFN is used to process information in the channel dimension. Compared to conventional Feed-forward Network (FN), CEFN improves the efficiency of the network optimization by using channel attention \cite{SENet}. CEFN can treat different channels unequally, making the network pay more attention to the channels with more critical information. Furthermore, our LKD Block can be considered a Transformer-like CNNs plug-in,  which uses the designed DLKCB to replace the MHSA in Transformer to enhance the performance, and uses the CEFN to replace the FN in CNNs to improve efficiency. Thus, LKD Block can be used as a plug-in module in both CNNs and ViTs for high-level and low-level computer vision tasks. Moreover, our DLKCB architecture has the same scale-up capability as the Transformer-based \cite{1}, \cite{ViT} methods, enabling our network to be more adaptable to devices with different computational performances than traditional CNNs. Detailed experimental results will demonstrate the efficiency of our LKD Block compared to other architecture.

\begin{figure}[t]
	\centering
	\includegraphics[width=0.95\columnwidth]{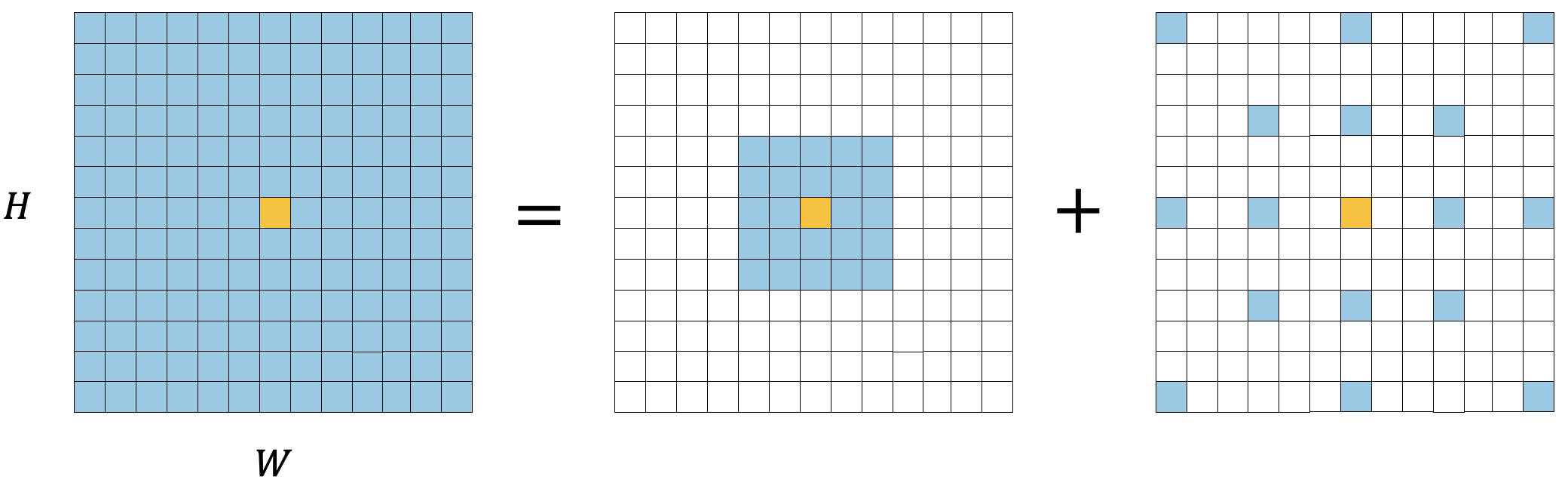} % Reduce the figure size so that it is slightly narrower than the column. Don't use precise values for figure width.This setup will avoid overfull boxes. 
	\caption{ Decomposition diagram of a large depth-wise convolution. The original large depth-wise convolution into a smaller depth-wise convolution and a depth-wise dilated convolution. The blue grids represent the location of the convolution kernel, and the yellow grid represents the center of convolution. The diagram shows that a $13\times13$ depth-wise convolution can be combined with a $5\times5$ depth-wise convolution and a $5\times5$ depth-wise dilated convolution with dilation rate 3. }
	\label{fig:decompose}
\end{figure}

\subsection{Decomposition Large Kernel Convolution Block}
A large receptive field increases the deep model’s ability to capture more structured information in the feature domain space \cite{3}, which is critical for image dehazing. The most popular approach is to increase the depth of the network by stacking several small convolutions \cite{2}, \cite{3}, \cite{4} (e.g., 3 × 3 convolutions) to increase the receptive field. However, this approach can increase the theoretical receptive field but is limited in increasing the effective receptive fields (ERFs) \cite{erfs}. Numerous works \cite{long2015fully}, \cite{peng2017large}, \cite{yu2017dilated} have demonstrated the critical role of ERFs in the performance of vision tasks. Recent work \cite{repLK} has shown that large depth-wise convolution can effectively increase ERFs. However, direct use of large depth-wise convolutions introduces massive computational overhead and parameters. To alleviate this issue, we propose Decomposition Large Kernel Convolution Block, as shown in Figure \ref{fig:flowchart} (b). The details of the decomposition of large kernel convolution are shown in Figure \ref{fig:decompose}, and the conventional large depth-wise convolution is decomposed into a smaller depth-wise convolution and a depth-wise dilated convolution. The equations of the decomposition of large depth-wise convolution of the number of parameters $P (K, d)$ and FLOPs $F (K, d)$ are expressed as follows:
\begin{eqnarray}
\label{eqn:params}
   P(K, d)&=&C (\lceil \frac{K}{d} \rceil ^ 2  \times C + (2d - 1)^2), \\
  F(K, d)&=&P(K, d) \times H \times W.   
\end{eqnarray}

Where $K$ denotes kernel size and $d$ is dilation rate. 

\begin{figure}[t]
	\centering
	\includegraphics[width=0.8\columnwidth]{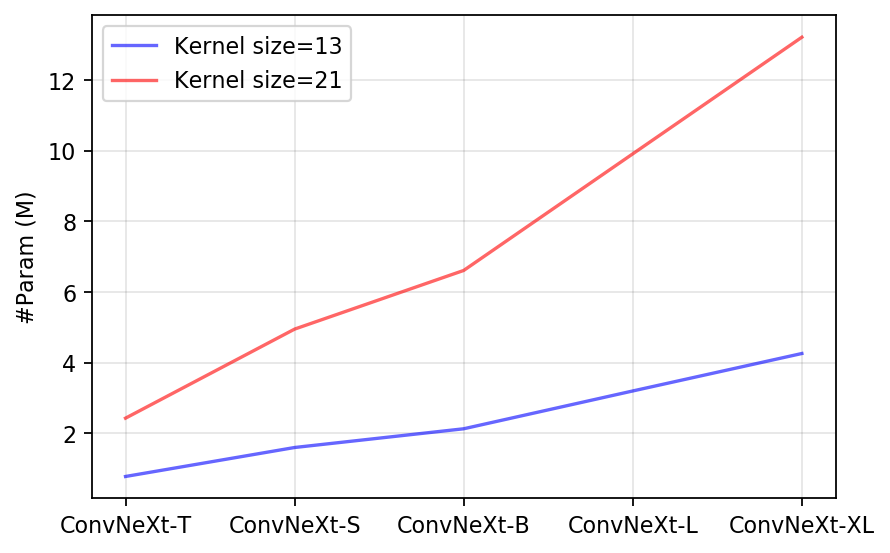} % Reduce the figure size so that it is slightly narrower than the column. Don't use precise values for figure width.This setup will avoid overfull boxes. 
	\caption{ For different kernel sizes, there are parameter differences between decomposition of depth-wise convolutions and depth-wise convolutions on the ConvNeXt. }
	\label{fig:cdbsd}
	
\end{figure}

\begin{figure}[t]
	\centering
	\includegraphics[width=1.0\columnwidth]{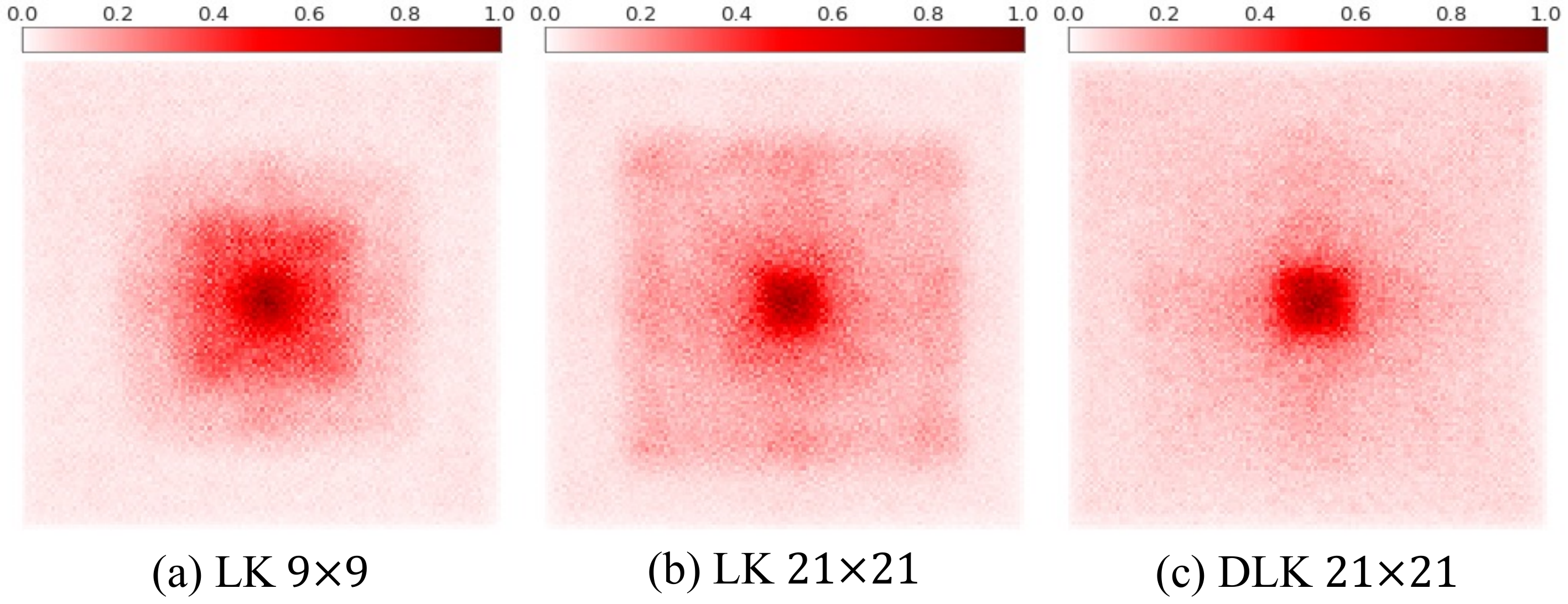}
	\caption{ The Effective Receptive Fields (ERFs) visualization results in different kernels. (a) $9\times9$ large kernel depth-wise convolution. (b) $21\times21$ large kernel depth-wise convolution. (c) $21\times21$ decomposition of large kernel depth-wise convolution. A more widely distributed red area indicates a larger ERF. Compared to LK $9\times9$ and LK $21\times21$, our DLK $21\times21$ obtains larger ERFs. }
	\label{fig:erfs}
\end{figure}

As shown in Figure \ref{fig:cdbsd}, we compare the number of parameters between the direct use of large depth-wise convolutions and the decomposition of large depth-wise convolutions on the ConvNeXt \cite{liu2022convnet}. We can observe that the parameter gap becomes more significant with the size of the network and the kernel, so we argue that our DLKCB will be more efficient in large networks. As shown in Figure \ref{fig:erfs}, we also find that decomposition of large kernel depth-wise convolution obtains larger ERFs in practice. The detail can be seen in the ablation study. 

\subsection{Channel Enhanced Feed-forward Network}
Some studies \cite{2}, \cite{SENet} have shown that different channel features have completely different weighting information. In other words, some channel features are not particularly important in network optimization. Therefore, if we treat these channels equally, we will put resources on that less critical information and affect the network's performance. For this purpose, we propose the CEFN, as shown in Figure \ref{fig:flowchart} (b). We incorporate channel attention into the conventional Feed-forward Network (FN), which allows the conventional FN to re-weight different channel features. Further, following \cite{van}, \cite{li2021convmlp}, we put a $3\times3$ depth-wise convolution into the conventional FN so that it can encode the information of neighboring pixel's positions on the feature space. The formula for CEFN is expressed as follows:
\begin{eqnarray}    \label{eqn:CEFN}
\hat{X}&=&X + norm(   \nonumber \\
&\;&norm(FN(X)\odot{CA(X)\cdot\alpha}),
\end{eqnarray}

Where $\odot{}$ denotes element-wise multiplication. $X\in{\mathbb R^{H\times{W\times{C}}}}$ and $\hat{X}\in{\mathbb R^{H\times{W\times{C}}}}$ are the input and output feature maps. $norm$ is the batch normalization. $\alpha$ is a learnable scaling parameter. $FN$ is the Feed-forward Network. $CA$ denotes channle attention, and its formula is expressed as follows:
\begin{eqnarray}    \label{eqn:CA}
CA(X)&=&\sigma(Linear(ReLU(\nonumber    \\
&\;&Linear(GAP(X))))),
\end{eqnarray}

Where $\sigma$ is the sigmoid function and $GAP$ denotes the global average pooling operation. Detailed experiments demonstrate that our CEFN is more effective than the FN applied to Transformers \cite{1}, \cite{dehazeformer}, \cite{ViT} and MLPs \cite{liu2021pay}.

\begin{table*}[t]
  \centering
  \begin{center}
    \renewcommand\arraystretch{1.25}
    {
      \begin{tabular}{c|cc|cc|cc}
        \toprule[1pt]
        \multirow{3}*{Methods}&\multicolumn{2}{c|}{ITS}&\multicolumn{2}{c|}{OTS} & \multicolumn{2}{c}{\multirow{2}{*}{Overhead}}\\
        \cline{2-5}
         &\multicolumn{2}{c|}{SOTS indoor}&\multicolumn{2}{c|}{SOTS outdoor} & \multicolumn{2}{c}{}\\
        \cline{2-7}
                                    & PSNR  & SSIM  & PSNR  & SSIM  & \#Param & FLOPs \\\hline
        DCP (TPAMI'10)              & 16.62 & 0.818 & 19.13 & 0.815 & -      & -      \\
        DehazeNet (TIP'16)          & 21.14 & 0.847& 22.46 & 0.8514& 0.008M  & 0.514G \\
        AOD-Net (ICCV'17)           & 19.06 & 0.850& 20.29 & 0.8765 & 0.0018M& 0.114G \\
        GFN (CVPR'18)               & 22.30 & 0.880 & 21.55 & 0.844 & 0.499M & 14.94G \\
        GridDehazeNet (ICCV'19)     & 32.16 & 0.984 & 30.86 & 0.982 & 0.956M & 18.77G \\
        MSBDN (CVPR'20)             & 33.67 & 0.985 & 33.48 & 0.982 & 31.35M & 24.44G \\
        PFDN (ECCV'20)              & 32.68 & 0.976 & -     & -     & 11.27M & 50.46G \\
        FFA-Net (AAAI'20)           & 36.39 & 0.989 & 33.57 & 0.984 & 4.456M & 287.53G \\
        AECR-Net (CVPR'21)          & 37.17 & 0.990 & -     & -     & 2.611M & 52.20G \\
        UDN (AAAI'22)               & \underline{38.62} & 0.991 & \underline{34.92} & \textbf{0.987} & 4.25M  & - \\
        Dehamer (CVPR'22)           & 36.36 & 0.988 & \textbf{35.18} & \underline{0.986} & 132.45M& 48.93G \\
        MAXIM (CVPR'22)             & 38.11 & 0.991 & 34.19 & 0.985 & 14.1M  & 216G \\
        \hline
        LKD-T (Ours)             & 34.77 & 0.987 & 33.33 & 0.980 & 0.343M  & 3.41G \\
        LKD-S (Ours)             & 37.04 & 0.991 & 33.97 & 0.982 & 0.634M  & 6.34G \\
        LKD-B (Ours)             & 38.57 & \underline{0.993} & 34.81 & 0.983 & 1.216M   & 12.20G \\
        LKD-L (Ours)             & \textbf{39.44} & \textbf{0.994} & 34.82 & 0.983 & 2.38M  & 23.93G \\
        \bottomrule[1pt]
      \end{tabular}
    }
  \end{center}
    \caption{
        Quantitative comparison of various dehazing methods trained on the ITS and OTS datasets in terms of PSNR, SSIM, number of the parameter (\#Param), and floating point operations (FLOPs). We use \textbf{bold} and \underline{underline }to indicate the highest and second highest results, respectively. The sign "-" denotes the number is unavailable. Note: FLOPs are measured on $256\times256$
        size images.}
    \label{tab3}
\end{table*}

\begin{figure*}[t]
	
	\centering
	\includegraphics[scale=0.45]{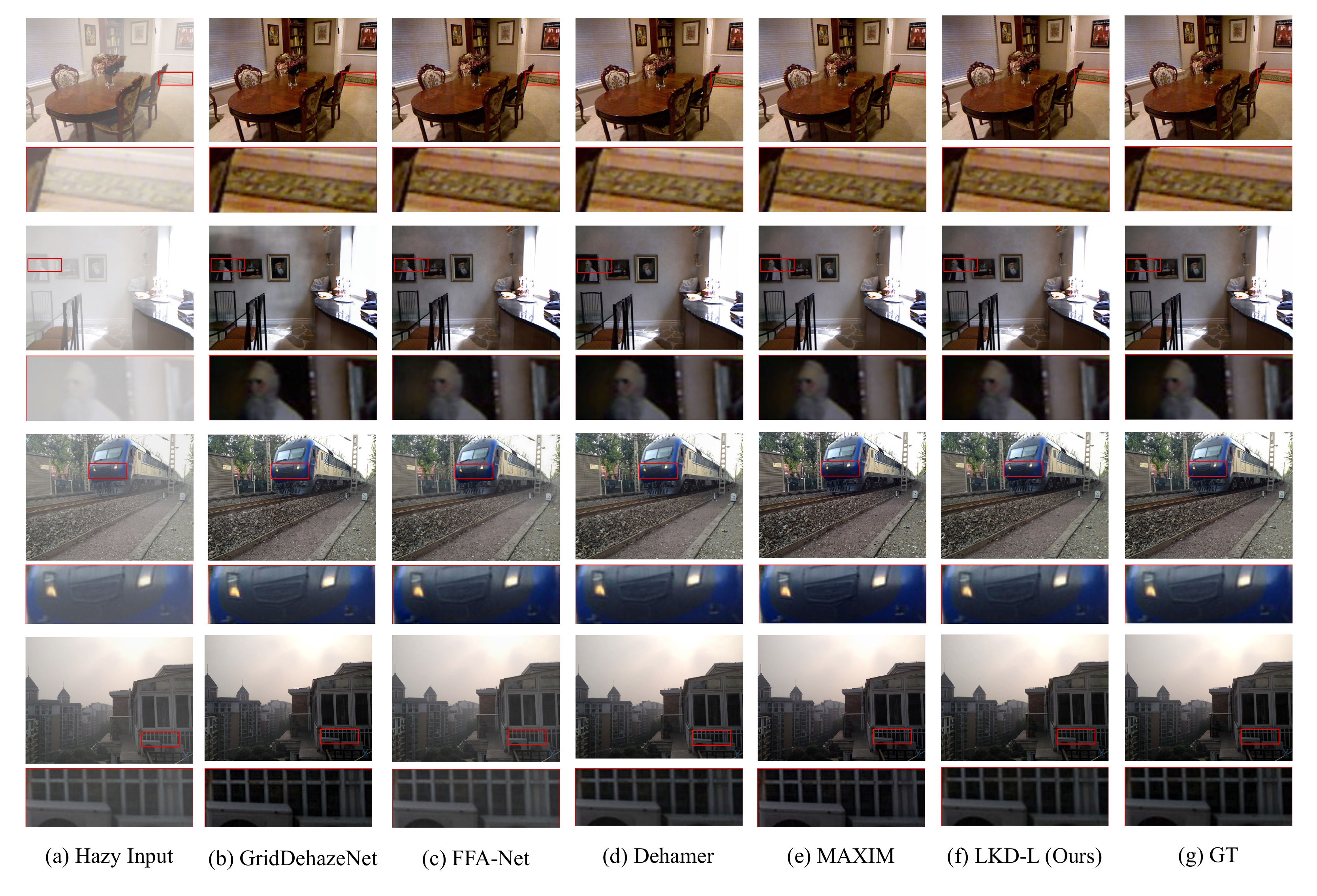} % Reduce the figure size so that it is slightly narrower than the column.
	\caption{ Qualitatively comparing image dehazing methods on SOTS set (zooming in for a better view). The first two rows are SOTS indoor images, and the last two are outdoor images. The first column is the hazy input images, and the last is the corresponding ground truth. }
	\label{fig4}
	
\end{figure*}

\section{Experiments}
\subsection{Implementation Details}
We classify our LKD-Net into LKD-T, LKD-S, LKD-B, and LKD-L according to the number of their parameters and computational overhead, which correspond to tiny, small, basic, and large, respectively. Table  \ref{tab2} lists the configuration details of these variants. All models are implemented with the PyTorch 1.10.1 on two NVIDIA TITAN Xp GPUs. The AdamW \cite{DBLP:conf/iclr/LoshchilovH19} optimizer is utilized to optimize our LKD-Net with exponential decay rates $\beta_1$ and $\beta_2$ equals to 0.9 and 0.999, respectively. The initial learning rate is set to 0.0002, and the cosine annealing strategy is used to adjust the learning rate. The batch size is set to 16, and the patch size is set to $256\times256$ with a random crop. We only use $L_1$ loss to optimize our LKD-Net. We decompose the $21\times21$ convolution by default, which is proven to have the best parameter-performance trade-off in the work \cite{van}.

\begin{table}[ht]
    \centering
    \setlength\tabcolsep{3pt}
    \begin{tabular}{cccc}
        \toprule[1pt]
        Model & Num. Blocks & Embedding Dims & MLP Ratio    \\
        \hline
        LKD-T & [1, 1, 2, 1, 1] & [24, 48, 96, 48, 24] & [4, 4, 4, 4, 4] \\
        LKD-S & [2, 2, 4, 2, 2] & [24, 48, 96, 48, 24] & [4, 4, 4, 4, 4] \\
        LKD-B & [4, 4, 8, 4, 4] & [24, 48, 96, 48, 24] & [4, 4, 4, 4, 4] \\
        LKD-L & [8, 8, 16, 8, 8] & [24, 48, 96, 48, 24] & [4, 4, 4, 4, 4] \\
        \bottomrule[1pt]
    \end{tabular}
    
    \caption{The configuration detail of LKD-Net.}
    \label{tab2}
\end{table}

\subsection{Datasets and Evaluation Metrics}
Our experiments are evaluated on RESIDE dataset \cite{RESIDE}. RESIDE is one of the most commonly used datasets for single image dehazing tasks, which contains five subsets: Indoor Training Set (ITS), Outdoor Training Sets (OTS), Synthetic Objective Testing Set (SOTS), Real World task-driven Testing Set (RTTS) and Hybrid Subjective Testing Set (HSTS). The ITS subset contains 13,990 image pairs, and the OTS subset contains 313,950 image pairs. The SOTS contains 500 indoor and 500 outdoor test image pairs. Following the objective evaluation protocol\cite{2}, \cite{3}, \cite{dehazeformer}, Our LKD-Net is trained on ITS for 300 epochs and OTS for 30 epochs, respectively, and evaluated on the SOTS subset. Meanwhile, the Peak Signal Noise Ratio (PSNR) and the Structure Similarity index (SSIM) are used to evaluate the performance of our LKD-Net and the compared state-of-the-art methods.

\subsection{Results on RESIDE Dataset}
We quantitatively compare the performance of our LKD-Net and the state-of-the-art image dehazing methods, including the DCP \cite{DCP}, DehazeNet \cite{DehazeNet}, AOD-Net \cite{4}, GFN \cite{ren2018gated}, and GridDehazeNet \cite{liu2019griddehazenet}, MSBDN \cite{dong2020multi}, PFDN \cite{dong2020physics}, FFA-Net \cite{2}, AECR-Net \cite{3}, UDN \cite{UDN}, Dehamer \cite{guo2022image}, MAXIM \cite{tu2022maxim}. The comparison results are shown in Table \ref{tab3}. It can be seen that our LKD-L outperforms all methods on the SOTS indoor dataset. It outperforms the previous best method UDN by 0.86 dB in the PSNR evaluation metric with only 56\% of the number of parameters. In particular, compared with the Transformer-based method Dehamer, LKD-L outperforms Dehamer by 3.08 dB in the PSNR evaluation metric with only 1.79\% of the number of parameters and 48.9\% of the FLOPs, which demonstrates that the Transformer-based method may not the optimal option for low-level computer vision tasks. In addition, benefiting from the fantastic architecture of LKD-Block, all variants of LKD-Net have achieved good performance, so we believe LKD-Net is a scalable method, which makes it adaptable to a wide range of devices with different performances. On the SOTS outdoor dataset, our LKD-L does not perform as well as UDN and Dehamer, but the difference is not particularly significant (the maximum difference is only 0.36dB). However, they have several times the number of parameters of LKD-L, and we argue that our network achieves a better parameter-performance trade-off.

We also qualitatively compare our LKD-Net with the state-of-the-art image dehazing methods, and the visualization results are shown in Figure \ref{fig4}. We can see that GridDehzeNet and FFA-Net cannot successfully remove the hazy from the images. Dehamer performs well in outdoor scenes. However, it also suffers from color distortion in indoor scenes (e.g., the background color of the characters changes in the second row). In contrast, MAXIM performs well in indoor scenes. However, color oversaturation occurs in outdoor scenes (e.g., oversaturation of train lights in the third column). Compared to them, the recovered images from our LKD-Net are significantly closer to Ground Truth.

\subsection{Ablation Study}
Ablation studies are conducted to demonstrate the effectiveness of each proposed component in LKD-Net, and the minimal model LKD-T is used for the ablation analysis. We first construct a \textbf{Base} network as our baseline, which is implemented by replacing $21\times21$ decomposition of depth-wise convolutions with $7\times7$ depth-wise convolutions, replacing CEFN with the regular Feed-forward Network (FN), replacing SK Fusion with concatenation, and replacing SR with global residual learning. Subsequently, we replace different modules into base network construct four different variants: (1) \textbf{Base+SF}: Replace concatenation in the \textbf{Base} with SK Fusion. (2) \textbf{Base+SF+SR}: Replace global residual learning in the \textbf{Base+SF} with Soft Reconstruction. (3) \textbf{Base+SF+SR+DLK}: Replace $7\times7$ depth-wise convolutions in the \textbf{Base+SF+SR} with $21\times21$ decomposition of depth-wise convolutions. (4) \textbf{Base+SF+SR+CEFN}: Replace the regular Feed-forward Network (FN) in the \textbf{Base+SF+SR} with CEFN. (5) \textbf{Ours}: Replace $7\times7$ depth-wise convolutions in the \textbf{Base+SF+SR+CEFN} with $21\times21$ decomposition of depth-wise convolutions. These models are trained on ITS dataset and evaluated on SOTS indoor set. The performance of these models is summarized in Table \ref{tab4}.

\begin{table}[ht]
    \centering
    \setlength\tabcolsep{4pt}
    \begin{tabular}{l|cccc}
        \toprule[1pt]
        Model & PSNR & SSIM & \#Param & FLOPs \\
        \cline{1-5}
        \textbf{Base}& 29.37 & 0.958 & \textbf{0.310M} & \textbf{2.98G}\\
        \textbf{Base+SF} & 31.74 & 0.977 & 0.314M & 3.06G \\
        \textbf{Base+SF+SR} & 31.83 & 0.978 & 0.315M & 3.07G \\
        \textbf{Base+SF+SR+DLK} & 33.78 & 0.985 & 0.337M & 3.40G \\
        \textbf{Base+SF+SR+CEFN} & 33.38 & 0.983 & 0.334M & 3.27G \\
        \cline{1-5}
        \textbf{Ours} & \textbf{34.77} & \textbf{0.987} & 0.343M & 3.41G \\
        \bottomrule[1pt]
    \end{tabular}
    \caption{Ablation study on LKD-T with different architectures.}
    \label{tab4}
\end{table}

\textbf{Effectiveness of DLK.} Compared to Base+SF+SR, the DLK can significantly increase the PSNR by 1.95dB and SSIM by 0.007, while only introducing 0.022M \#Param and 0.33G FLOPs. Furthermore, we also performed ablation experiments with other conventional large depth-wise convolutions to demonstrate the advantages of our DLK. Table \ref{tab5} shows that our DLK achieves better performance using less computational overhead and parameters than $9\times9$ depth-wise convolution and $21\times21$ depth-wise convolution. We argue that the main reason for the high effectiveness of DLK is that it can capture larger ERFs compared to conventional large depth-wise convolutions. To support our viewpoint, we use the public tool (code is available at \cite{repLK}) to visualize the ERF of the feature map centroids of the LKD-Net $L_3$ output. As shown in Figure \ref{fig:erfs}, we can observe that the distribution of red dots in DLK $21\times21$ is larger and more widely than LK $9\times9$ and LK $21\times21$, which indicates that DLK $21\times21$ obtains larger ERF indeed. We also present quantitative analysis in Table \ref{tab:erfs}, where $t$ stands a threshold. For example, if $t=20\%$ and $r=4.4\%$ that means 20\% of the pixel contributions reside in 4.4\% total pixel area. We can see that DLK $21\times21$ has a smoother distribution of high contributing pixels compared to LK $9\times9$ and LK $21\times21$, which proves our viewpoint that the main reason for the high effectiveness of DLK is that it can effectively capture larger ERFs.

\begin{table}[ht]
    \centering
    \begin{tabular}{l|cccc}
        \toprule[1pt]
        Kernel size & PSNR & SSIM & \#Param & FLOPs \\
        \cline{1-5}
        LK $9\times9$ & 33.95 & 0.985 & 0.3452M & 3.45G\\
        LK $21\times21$ & 34.31 & 0.987 & 0.4662M & 5.430G \\
        DLK $21\times21$ & \textbf{34.77} & \textbf{0.987} & \textbf{0.343M} & \textbf{3.41G}\\
        \bottomrule[1pt]
    \end{tabular}
    
    \caption{ Comparisons of different kernel size conventional depth-wise convolutions. }
    \label{tab5}
\end{table}

\textbf{Effectiveness of CEFN.} Compared to Base+SF+SR, CEFN significantly improves performance with a 1.55 dB increase in PSNR and a 0.005 increase in SSIM and only introduces 0.019M \#Param and 0.2G FLOPs. We believe that the main reason for the high effectiveness of CEFN is that the channel attention mechanism \cite{SENet} allows CEFN to focus more on the channels with important information.

\begin{table}[ht]
    \centering
    \setlength\tabcolsep{4pt}
    \begin{tabular}{l|cccc}
        \toprule[1pt]
        Kernel size & $t=20\%$ & $t=30\%$ & $t=50\%$ & $t=99\%$  \\
        \cline{1-5}
        LK $9\times9$ & 4.4\% & 7.5\% & 17.1\% & 95.3\% \\
        LK $21\times21$ & 6.6\% & 12.4\% & 27.4\% & 95.4\% \\
        DLK $21\times21$ & 6.6\% & 13.5\% & 30.8\% & 98.4\% \\
        \bottomrule[1pt]
    \end{tabular}
    
    \caption{ Quantitative analysis on the ERF with the high-contribution area ratio $r$. A larger $r$ indicates a smoother distribution of high-contribution pixels, hence larger ERF. }
    \label{tab:erfs}
\end{table}

\section{Conclusion}
This paper proposes a novel LKD-Net for high-performance single image dehazing. The designed DLKCB can effectively capture ERFs and model long-range information, and the designed CEFN can effectively enhance channel dimension features in FN. Evaluation results show that LKD-Net outperforms the state-of-the-art and dramatically outperforms the Transformer-based method Dehamer. Thus, we argue that our LKD-Net is an effective and universal end-to-end image restore method, and can be used for video dehazing and other low-level computer vision tasks such as image denoising, rain removal, deblurring, super-resolution, etc. Moreover, the decomposition of depth-wise convolutions in DLKCB may be used in CNNs and ViTs to enhance the performance of both low-level and high-level vision tasks.

% Use \bibliography{yourbibfile} instead or the References section will not appear in your paper
\bibliography{aaai23}

\end{document}